\documentclass{article}

\PassOptionsToPackage{numbers}{natbib}

 \usepackage[dblblindworkshop, final]{neurips_2025}
 \usepackage{amsmath}
\usepackage{booktabs}
\usepackage{graphicx}
\usepackage{placeins} 
\usepackage{float}    

\workshoptitle{First Workshop on CogInterp: Interpreting Cognition in Deep Learning Models}

\usepackage[utf8]{inputenc} 
\usepackage[T1]{fontenc}    
\usepackage{hyperref}       
\usepackage{url}            
\usepackage{booktabs}       
\usepackage{amsfonts}       
\usepackage{nicefrac}       
\usepackage{microtype}      
\usepackage{xcolor}         

\usepackage{subcaption}

\newcommand{\saliencygrid}[9]{%
  \setlength{\tabcolsep}{1pt}%
  \renewcommand{\arraystretch}{0}%
  \begin{tabular}{@{}ccc@{}}
  \includegraphics[width=0.32\linewidth]{#1} &
  \includegraphics[width=0.32\linewidth]{#2} &
  \includegraphics[width=0.32\linewidth]{#3} \\
  \includegraphics[width=0.32\linewidth]{#4} &
  \includegraphics[width=0.32\linewidth]{#5} &
  \includegraphics[width=0.32\linewidth]{#6} \\
  \includegraphics[width=0.32\linewidth]{#7} &
  \includegraphics[width=0.32\linewidth]{#8} &
  \includegraphics[width=0.32\linewidth]{#9} \\
  \end{tabular}%
}

\title{Learning to Look: Cognitive Attention Alignment with Vision-Language Models}

\author{%
  Ryan L. Yang\thanks{Equal contribution.}\\
  Brown University\\
  Providence, RI   \\
  \texttt{ryan\_l\_yang@brown.edu} \\
  \And
  Dipkamal Bhusal\footnotemark[1]\\
  Rochester Institute of Technology\\
  Rochester, NY  \\
  \texttt{db1702@rit.edu } \\
  \And
  Nidhi Rastogi \\
  Rochester Institute of Technology\\
  Rochester, NY \\
  \texttt{nxrvse@rit.edu} \\
}

\begin{document}

\maketitle

\begin{abstract}
Convolutional Neural Networks (CNNs) frequently “cheat” by exploiting superficial correlations, raising concerns about whether they make predictions for the right reasons. Inspired by cognitive science, which highlights the role of attention in robust human perception, recent methods have sought to guide model attention using concept-based supervision and explanation regularization. However, these techniques depend on labor-intensive, expert-provided annotations, limiting their scalability. We propose a scalable framework that leverages vision-language models to automatically generate semantic attention maps using natural language prompts. By introducing an auxiliary loss that aligns CNN attention with these language-guided maps, our approach promotes more reliable and cognitively plausible decision-making without manual annotation. Experiments on challenging datasets, ColoredMNIST and DecoyMNIST, show that our method achieves state-of-the-art performance on ColorMNIST and remains competitive with annotation-heavy baselines on DecoyMNIST, demonstrating improved generalization, reduced shortcut reliance, and model attention that better reflects human intuition. Our code is available at \url{https://github.com/ryanlyang/LearningToLook/}. 
\end{abstract}

\section{Introduction}

Despite their impressive performance, Convolutional Neural Networks (CNNs) tend to ``cheat” during learning, exploiting superficial correlations and spurious features in data rather than acquiring robust, generalizable representations~\cite{geirhos2020shortcut}. This shortcut learning raises a critical question for building reliable AI systems: \textit{Are models right for the right reasons?}.

Cognitive science offers a rich tradition of studying not only what intelligence can do, but how it achieves its capabilities. Human perception, for example, is guided by attention mechanisms that flexibly allocate cognitive resources to task-relevant features. One of the simplest examples is how we identify objects by their distinctive shapes, patterns or colors, ignoring irrelevant distractions. These mechanisms support robust generalization and interpretable reasoning. Inspired by such insights, recent efforts have influenced the strategies for attention and feature selection in neural networks~\cite{ross2017right, rieger2020interpretations, gupta2023concept}. 

One of the popular approaches is to guide models toward human-meaningful, task-relevant regions when making decisions. Prior work has introduced concept-based supervision~\cite{gupta2023concept} and explanation regularization techniques~\cite{ross2017right, rieger2020interpretations}, where models are encouraged to align their saliency or concept activations with expert-provided ground-truth annotations through auxiliary losses. However, these approaches require manual collection of ground truth for saliency or, dataset for concepts with deep domain expertise, and can introduce annotation biases, restricting their scalability and applicability.

In this work, we present a scalable framework that leverages advances in vision-language models to automate cognitively meaningful attention supervision. Specifically, we utilize WeCLIP+~\cite{zhang2025frozen} to generate attention maps for arbitrary visual concepts using natural language prompts, serving as ``teacher” signals grounded in semantics. During training, we introduce an auxiliary loss that aligns the model’s attention with these language-guided reference maps, steering CNNs toward more reliable, and cognitively plausible decision-making, without the bottleneck of manual annotation. 

We seek to explore the following research questions: \textit{Can we use language and vision to instill cognitively motivated inductive biases in neural networks?} \textit{Does this reduce shortcut reliance and improve robustness or generalization?} We investigate these questions on challenging classification tasks such as ColoredMNIST~\cite{li2019repair} and DecoyMNIST~\cite{erion2021improving}. Our method achieves state-of-the-art results on ColoredMNIST and remains competitive with annotation-intensive baselines on DecoyMNIST, despite relying only on automatically generated pseudo-maps.

\section{Related work}
Concept Distillation~\cite{gupta2023concept} utilizes the concept activation vectors (CAVs) framework from TCAV~\cite{kim2018interpretability} by introducing a concept loss to fine-tune models for reducing bias and improving alignment with human-understood concepts. However, such methods rely on manually defined and collected concept samples, which are costly to annotate and may introduce biases reflective of the provided dataset. Another approach, CDEP~\cite{rieger2020interpretations}, enables the integration of domain knowledge by penalizing models whose explanations do not align with expert-identified, task-relevant features. However, it still requires extensive domain expertise and human annotation to construct ground-truth explanations, limiting its scalability. Similarly, Right for the Right Reasons~\cite{ross2017right} introduces an input gradient regularization technique, where model training is guided by selectively penalizing input gradients corresponding to features identified (by experts) as irrelevant or spurious. While effective in encouraging models to base decisions on the ``right” features, this approach also depends on detailed, expert-provided annotations to specify which features should or should not influence model predictions.

\section{Methodology}

Our proposed framework consists of two principal stages: (1) generating class and concept-specific attention maps using a vision-language model, and (2) training a CNN with an auxiliary loss that aligns its attention with these automatically generated maps. In this section, we formally introduce our notation and describe each component in detail.

\subsection{Preliminaries and Notation}

Let $\mathcal{D} = \{(x_i, y_i)\}_{i=1}^N$ be a dataset of $N$ images $x_i \in \mathbb{R}^{H \times W \times 3}$ with corresponding class labels $y_i \in \mathcal{C}$, where $\mathcal{C}$ denotes the set of all possible classes. We denote our CNN model as $f_\theta$, parameterized by weights $\theta$, which outputs logits $f_\theta(x)$ for input $x$.

For an input $x$ and class $c$, a \emph{saliency map} $S_\theta(x, c) \in [0, 1]^{H \times W}$ visualizes the importance of each pixel in $x$ for predicting class $c$, typically obtained using a method such as IG \cite{sundararajan2017axiomatic} or Class Activation Mapping (CAM)\cite{zhou2016learning}. Similarly, $M_{\mathrm{VL}}(x, c) \in [0, 1]^{H \times W}$ denotes an attention map for concept $c$ in $x$ generated by a vision-language model.

\subsection{Automatic Attention Map Generation}
To obtain supervision signals without manual annotation, we employ WeCLIP+\cite{zhang2025frozen}, a state-of-the-art vision-language model, to generate class-specific attention maps. For each $(x_i, y_i)$, we construct a natural language prompt $t_{y_i}$ (e.g., ``a photo of a digit'') that matches the class semantics. Optionally, we also provide background or distractor prompts to help the model distinguish target concepts from context. In Appendix~\ref{appendix:prompts}, we discuss the prompts for our targeted classification tasks. 

WeCLIP+ computes an affinity map $M_{\mathrm{VL}}(x_i, y_i)$ that highlights image regions associated with the semantic concept $y_i$ and the natural language prompt. These maps are automatically generated for all images in the training and validation sets and used as pseudo ground-truth for attention alignment.

\paragraph{Attention Map Post-processing.}
To improve alignment with desired inductive biases, we optionally refine the raw attention maps. For instance, morphological dilation with a structuring element of radius $r$ can be applied to ensure that the entire object is covered, or edge detection (e.g., Canny operator followed by dilation \cite{canny2009computational}) can be used to focus the model on object boundaries. Such preprocessing is task-dependent and controlled via simple hyperparameters. However, we observed that the unmodified attention maps produced by WeCLIP+ often perform well enough on their own.

\subsection{Attention-Aligned CNN Training}

Given the dataset $\mathcal{D}$ and the set of generated attention maps $\{M_{\mathrm{VL}}(x_i, y_i)\}$, we train the CNN $f_\theta$ to simultaneously minimize classification error and encourage its internal attention to match the WeCLIP+ pseudo-masks.

\paragraph{Classification Loss.}
For each mini-batch, the standard cross-entropy loss is computed as:
\begin{equation}
    \mathcal{L}_{\mathrm{CE}} = -\frac{1}{B} \sum_{i=1}^B \log p_\theta(y_i \mid x_i),
\end{equation}
where $B$ is the batch size and $p_\theta(y_i \mid x_i)$ is the softmax probability of the correct class.

\paragraph{Attention Alignment Loss.}
For each sample, we compute a normalized saliency map $S_\theta(x_i, y_i)$ for the true label $y_i$ using class activation mapping (CAM). We then minimize the Kullback--Leibler (KL) divergence between this map and the WeCLIP+ attention map, both normalized so that $\sum_{h,w} S_\theta(x_i, y_i)[h, w] = 1$ and similarly for $M_{\mathrm{VL}}(x_i, y_i)$:
\begin{equation}
    \mathcal{L}_{\mathrm{attn}} = \frac{1}{B} \sum_{i=1}^B \mathrm{KL}\big(S_\theta(x_i, y_i) \;\|\; M_{\mathrm{VL}}(x_i, y_i)\big).
\end{equation}

The model is optimized with a weighted combination of classification and attention alignment losses:
\begin{equation}\label{eqn:newobject}
    \mathcal{L} = \mathcal{L}_{\mathrm{CE}} + \lambda\, \mathcal{L}_{\mathrm{attn}},
\end{equation}
where $\lambda > 0$ determines the strength of the attention supervision. 
We train with a two-phase schedule: for the first $E_{\mathrm{attn}}$ epochs we optimize only $\mathcal{L}_{\mathrm{attn}}$ to “learn to look”; at $E_{\mathrm{attn}}$ we reset the optimizer/scheduler and thereafter minimize the joint objective in Eqn.~\ref{eqn:newobject}.

\vspace{-1.5mm}
\section{Experiments}

We empirically evaluate our language-guided attention alignment framework on two challenging biased classification benchmarks: ColorMNIST~\cite{li2019repair} and DecoyMNIST~\cite{erion2021improving}. These datasets are designed to test a model’s reliance on spurious correlations and its ability to generalize beyond shortcut cues.

\subsection{Setup}
\paragraph{Datasets.}  \textbf{ColorMNIST} \cite{li2019repair} is a variant of MNIST~\cite{lecun2002gradient} in which each digit class is assigned a unique color in the training set. At test time, the color mapping is reversed, forcing models to rely on digit shape rather than color for accurate classification. In \textbf{DecoyMNIST}~\cite{erion2021improving}, MNIST digits are augmented by adding class-indicative gray patches to the image boundary. These “decoy” patches create a spurious association between digit class and patch location or intensity.

\paragraph{Baselines and Comparison.}
We compare our approach (\textbf{Ours}) to a baseline CNN (\textbf{Base}), concept distillation based supervision (CDBS)~\cite{gupta2023concept}, and explanation regularization techniques, right-for-right-reasons (RRR)~\cite{ross2017right} and penalizing explanations (CDEP)~\cite{rieger2020interpretations}. 

\paragraph{Training Details.}
We train a LeNet~\cite{lecun2002gradient} with SGD (momentum $0.98$, weight decay $10^{-4}$), batch size $32$, for $30$ epochs. The only dataset-specific hyperparameter is the initial learning rate: $10^{-3}$ for ColorMNIST and $10^{-2}$ for DecoyMNIST, decayed by a factor of $0.1$ every 7 epochs; all other settings are identical.

Training runs in two phases separated by the \emph{Attention epoch} $E_{\mathrm{attn}}$. 
Phase 1 (“learn to look”): optimize only $\mathcal{L}_{\mathrm{attn}}$ (KL between the model CAM and the WeCLIP\,+ pseudo-map). 
At $E_{\mathrm{attn}}$ we reset the optimizer and scheduler (zero momentum, restart LR) while keeping network weights unchanged. 
Phase 2: minimize $\mathcal{L}=\mathcal{L}_{\mathrm{CE}}+\lambda\,\mathcal{L}_{\mathrm{attn}}$ with a ramp on $\lambda$: for each epoch $e\ge E_{\mathrm{attn}}$, $\lambda_{e+1}=\lambda_{e}+0.1\,\lambda_{0}$ (with $\lambda_{E_{\mathrm{attn}}}=\lambda_{0}$) to keep attention prioritized.

We tune $\lambda$ and $E_{\mathrm{attn}}$ on a validation set using a composite metric we call \emph{Optim Value}:
\[
\text{Optim Value} \;=\; \text{ValAcc} \times \bigl(1 - \mathcal{L}_{\mathrm{attn}}\bigr),
\]
favoring configurations that jointly increase accuracy and decrease attention divergence. 

See Appendix~\ref{app:hparam} for the full $(\lambda, E_{\mathrm{attn}})$ grid search and selection criterion; heatmaps are shown in Fig.~\ref{fig:hparam-heatmaps}. We select $\lambda{=}160,\,E_{\mathrm{attn}}{=}11$ for ColorMNIST and $\lambda{=}8,\,E_{\mathrm{attn}}{=}13$ for DecoyMNIST

\paragraph{Mask preprocessing (ColorMNIST only).}
To bias the model toward \emph{shape} rather than color on ColorMNIST, we preprocess the WeCLIP\,+ pseudo-masks before computing $\mathcal{L}_{\mathrm{attn}}$. Concretely, we first apply a small morphological dilation to ensure the digit is fully covered, then extract a thin boundary band from the dilated mask (edge detection followed by a light dilation). For DecoyMNIST, we use the raw pseudo-masks without preprocessing.

\subsection{Result}

\begin{table}[h]
\centering 
\caption{Test accuracy (\%) on biased benchmarks. Higher is better. mean $\pm$ s.d. over 5 random seeds.}

\label{tab:results}
\resizebox{0.70\textwidth}{!}{%
\begin{tabular}{@{}lccccc@{}}
\toprule
 & \textbf{Base} & \textbf{CDEP}\cite{rieger2020interpretations} & \textbf{RRR}~\cite{ross2017right} & \textbf{CDBS}~\cite{gupta2023concept} & \textbf{Ours} \\ \midrule
\textbf{ColoredMNIST} & 0.1 & 31.0 & 0.1 & 50.93 & 64.88 $\pm$ 2.85 \\
\textbf{DecoyMNIST} & 52.8 & 97.2 & 99.0 & 98.9 & 96.19 $\pm$ 0.35\\ \bottomrule
\end{tabular}%
}
\end{table}

Table~\ref{tab:results} reports test accuracy on the biased dataset benchmarks. On \textbf{ColoredMNIST}, the baseline CNN (\textbf{Base}) fails completely, achieving only $0.1\%$ accuracy, confirming it has learned to classify digits entirely by color shortcuts.  The explanation-regularization method \textbf{RRR}~\cite{ross2017right} also performs poorly in this setting, while textbf{CDEP}~\cite{rieger2020interpretations} improves generalization to $31.0\%$ accuracy, and the concept-distillation approach \textbf{CDBS}~\cite{gupta2023concept} reaches $50.93\%$. Our method achieves the best result with $64.88 \pm 2.85\%$, demonstrating that language-guided attention alignment can reduce shortcut reliance more effectively than annotation-heavy baselines. Qualitative maps in Appendix~\ref{appendix:saliencymaps} illustrate that our method shifts saliency from background color to digit shape.

On \textbf{DecoyMNIST}, the baseline attains $52.8\%$, again showing substantial reliance on the corner-patch shortcut. Manual supervision methods perform near-perfectly, with \textbf{RRR} reaching $99.0\%$, \textbf{CDEP} $97.2\%$, and \textbf{CDBS} $98.9\%$. Our method achieves $96.19 \pm 0.35\%$, slightly below the annotation-heavy approaches but still competitive, while requiring no human-provided saliency or concept labels. Qualitative results (Appendix~\ref{appendix:saliencymaps}) confirm that our model attends primarily to the digit body rather than patch artifacts. 
shifts.

\section{Limitations}
Our work has few limitations that open avenues for future research. First, attention maps are precomputed and stored, which can be memory-intensive. In future work, we plan to explore on-the-fly generation techniques to integrate attention guidance directly into training, further enhancing efficiency and applicability. Second, the current evaluation is restricted to relatively simple datasets (ColoredMNIST and DecoyMNIST), and future work will investigate performance on more complex, high-dimensional benchmarks to assess broader generalizability. Finally, reliance on a vision-language model as an attention teacher may introduce its own biases. Future work will investigate strategies to mitigate such risks, for example through debiasing techniques or the use of multiple teachers.

\section{Conclusion}
We presented a scalable, annotation-free framework that leverages language-driven attention maps from vision-language models to guide neural networks toward task-relevant, human-meaningful features. Empirically, our approach achieves state-of-the-art accuracy on ColorMNIST and remains competitive with annotation-intensive baselines on DecoyMNIST, despite requiring no human-provided saliency maps or concept sets. Our framework is backbone-agnostic, supporting a variety of CNNs and differentiable saliency techniques, and can be extended to architectures such as Vision Transformers by adapting the attribution mechanism. 

\section*{Acknowledgement}
R. Yang and N. Rastogi were partially supported by NSF Award \#2447631.

\bibliographystyle{plainnat}
\bibliography{neurips}

\appendix

\newpage

\section{Saliency Map Comparison on ColorMNIST and DecoyMNIST}\label{appendix:saliencymaps}

\begin{figure*}[t]
\centering
\captionsetup[subfigure]{justification=centering}

\begin{subfigure}[t]{0.32\textwidth}
\centering
\saliencygrid
  {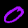} {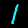} {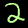}
  {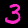} {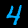} {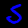}
  {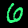} {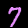} {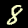}
\caption{ColorMNIST — originals}
\label{fig:color-orig}
\end{subfigure}\hfill
\begin{subfigure}[t]{0.32\textwidth}
\centering
\saliencygrid
  {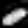} {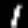} {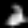}
  {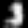} {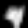} {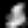}
  {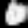} {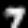} {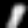}
\caption{ColorMNIST — saliency (Base)}
\label{fig:color-noattn}
\end{subfigure}\hfill
\begin{subfigure}[t]{0.32\textwidth}
\centering
\saliencygrid
  {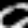} {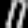} {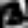}
  {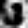} {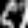} {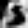}
  {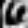} {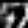} {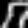}
\caption{ColorMNIST — saliency (Ours)}
\label{fig:color-attn}
\end{subfigure}

\vspace{0.6em}

\begin{subfigure}[t]{0.32\textwidth}
\centering
\saliencygrid
  {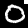} {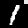} {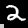}
  {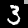} {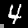} {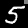}
  {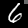}  {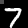} {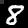}
\caption{DecoyMNIST — originals}
\label{fig:decoy-orig}
\end{subfigure}\hfill
\begin{subfigure}[t]{0.32\textwidth}
\centering
\saliencygrid
  {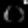} {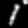} {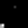}
  {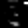} {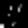} {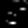}
  {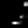}  {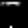} {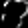}
\caption{DecoyMNIST — saliency (Base)}
\label{fig:decoy-noattn}
\end{subfigure}\hfill
\begin{subfigure}[t]{0.32\textwidth}
\centering
\saliencygrid
  {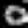} {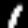} {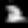}
  {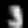} {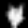} {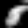}
  {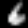} {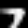} {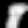}
\caption{DecoyMNIST — saliency (Ours)}
\label{fig:decoy-attn}
\end{subfigure}

\caption{\textbf{Qualitative comparison by dataset.} Each row shows 3×3 grids of \emph{(left)} original inputs, \emph{(middle)} saliency without attention alignment, and \emph{(right)} saliency with attention alignment. Brighter regions indicate higher saliency (e.g., CAM).}
\label{fig:saliency_triptych}
\end{figure*}

Figure~\ref{fig:saliency_triptych} demonstrates the saliency comparison using CAM before and after attention alignment. 

\FloatBarrier


\section{Prompts for Targeted Classification Tasks}
\label{appendix:prompts}

We use a single foreground class (\texttt{digit}) for both datasets to emphasize shape over spurious cues. Prompts are short, generic, and stable across images.

\subsection{ColoredMNIST}
\textbf{Foreground (class) prompts.}
\begin{itemize}
  \item \texttt{digit}
\end{itemize}

\textbf{Background categories.}
\begin{itemize}
  \item \texttt{Background}, \texttt{dark}, \texttt{black}
\end{itemize}

\subsection{DecoyMNIST}
\textbf{Foreground (class) prompts.}
\begin{itemize}
  \item \texttt{digit}
\end{itemize}

\textbf{Background categories.}
\begin{itemize}
  \item \texttt{Background}, \texttt{dark}, \texttt{black}, \texttt{corner}, \texttt{patch}, \texttt{box}, \texttt{corner patch}
\end{itemize}

\noindent\textit{Rationale.} We use a single neutral noun (\texttt{digit}) to keep the prompt stable and avoid leaking color or style cues, while the background tokens name the dominant nuisances in each dataset: \texttt{Background}, \texttt{dark}, and \texttt{black} capture canvas/intensity, and for DecoyMNIST \texttt{corner}, \texttt{patch}, \texttt{box}, \texttt{corner patch} explicitly describe the spurious square so WeCLIP+ separates it from the foreground.

\section{Hyperparameter Search}\label{app:hparam}

\FloatBarrier 

\begin{figure}[!htbp] 
  \centering
  \begin{subfigure}[t]{0.48\linewidth}
    \centering
    \includegraphics[width=\linewidth]{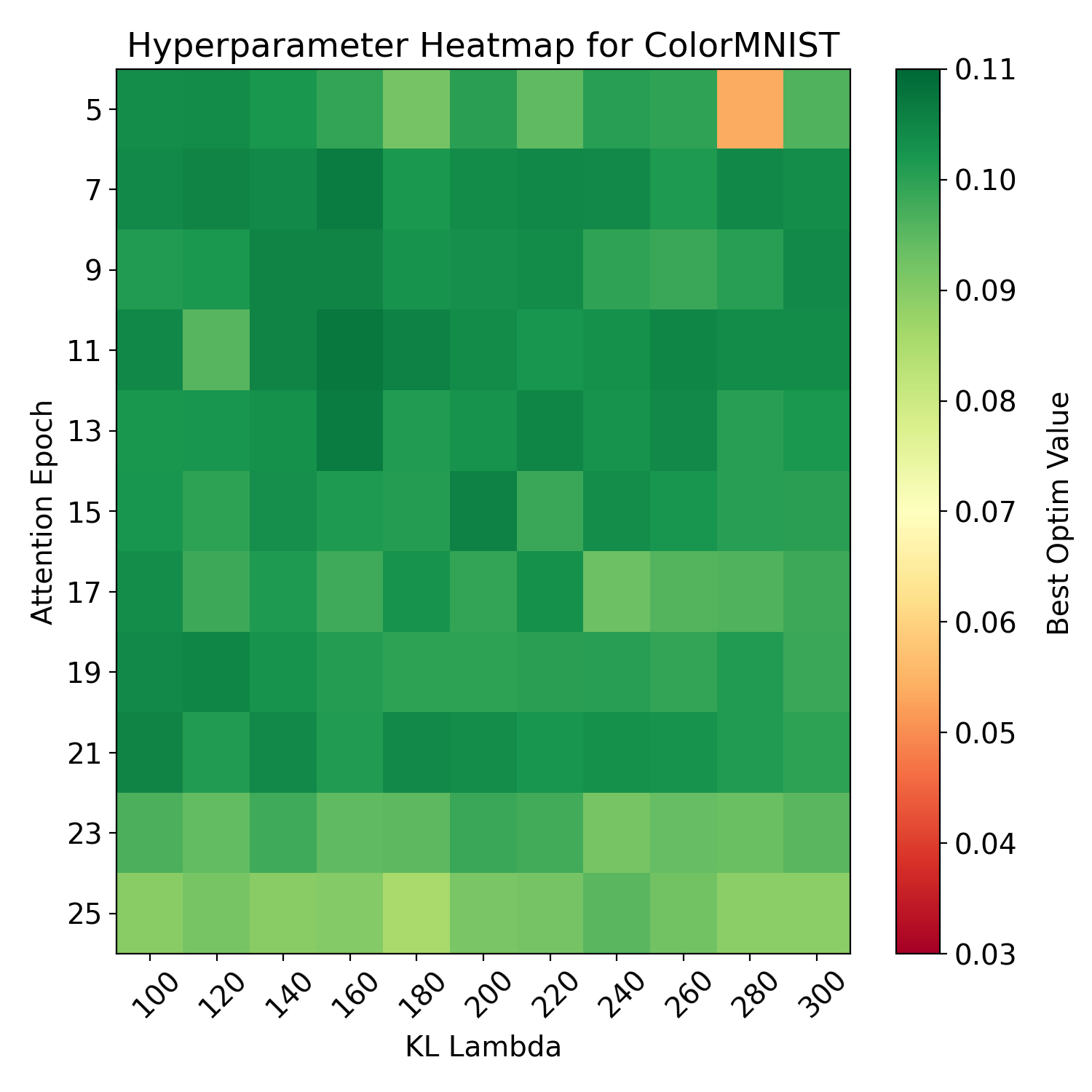}
    \caption{ColorMNIST: Optim Value across $(\lambda, E_{\mathrm{attn}})$.}
    \label{fig:hparam-color}
  \end{subfigure}\hfill
  \begin{subfigure}[t]{0.48\linewidth}
    \centering
    \includegraphics[width=\linewidth]{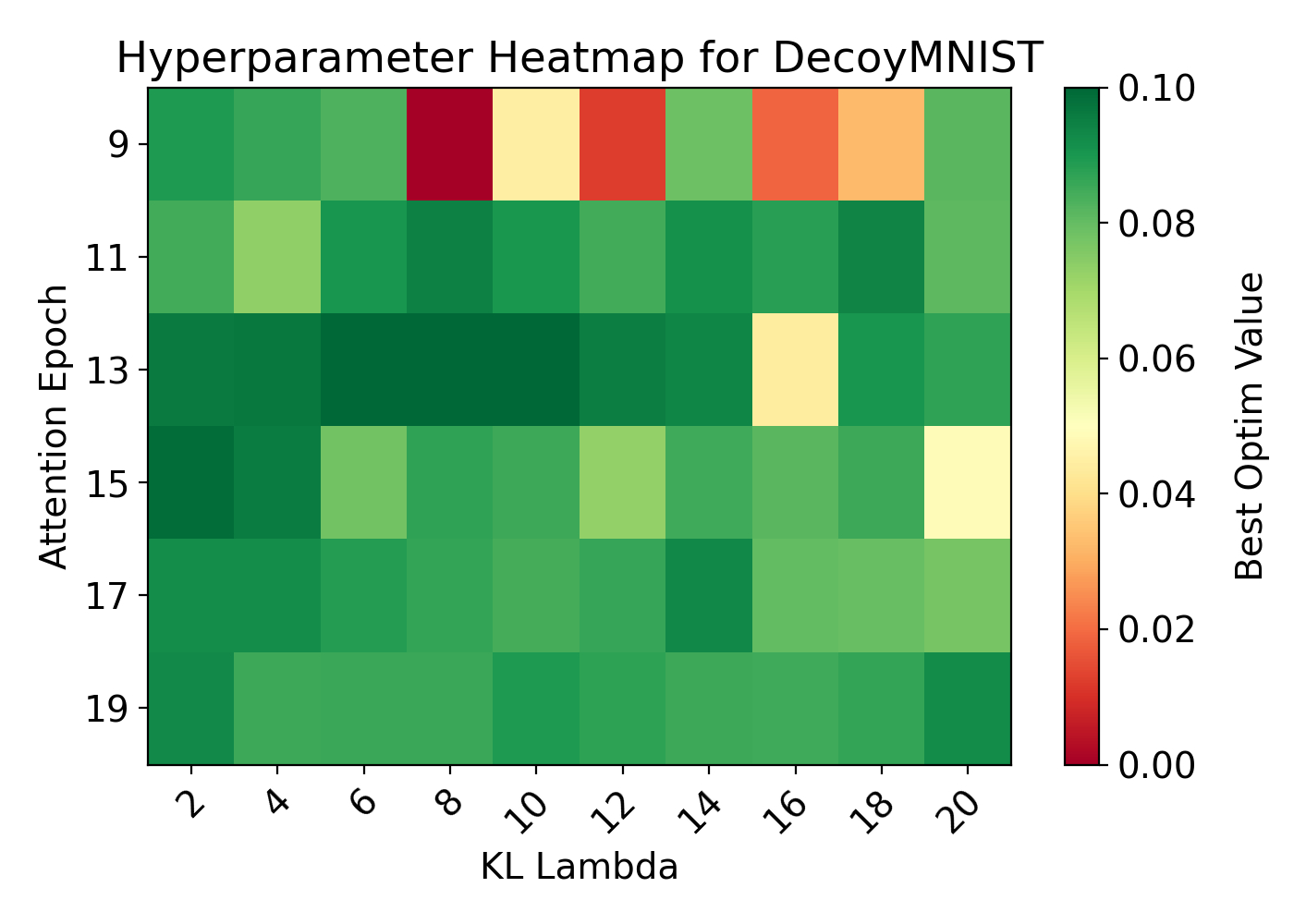}
    \caption{DecoyMNIST: Optim Value across $(\lambda, E_{\mathrm{attn}})$.}
    \label{fig:hparam-decoy}
  \end{subfigure}
  \caption{\textbf{Hyperparameter heatmaps.} Each cell shows the best \emph{Optim Value} during training (higher is better).}
  \label{fig:hparam-heatmaps}
\end{figure}

\FloatBarrier

We perform a grid search over the KL weight $\lambda$ and the Attention epoch $E_{\mathrm{attn}}$ (the epoch at which training switches from pure attention alignment to the combined objective in Eqn.~\ref{eqn:newobject}). Each cell reports the \emph{Optim Value} defined as
\[
\text{Optim Value} = \text{ValAcc} \times \bigl(1 - \mathcal{L}_{\mathrm{attn}}\bigr),
\]
so \emph{higher is better}. Training settings match the main text (SGD, $64$ batch, $30$ epochs); the initial learning rate is $10^{-3}$ for ColorMNIST and $10^{-2}$ for DecoyMNIST. $\mathcal{L}$ can be larger than 1, having a val optim number close to zero might still get decent accuracy on the test set, this is just a metric for optimizing hyperparameters

\paragraph{Note on the selection metric.}
Because $\mathcal{L}_{\mathrm{attn}}$ is a KL divergence, it is non-negative and \emph{unbounded}; in particular, $\mathcal{L}_{\mathrm{attn}}>1$ can occur. Consequently $1-\mathcal{L}_{\mathrm{attn}}$ may be small or even negative, so an \emph{Optim Value} near zero does not imply poor validation or test accuracy. We use this quantity solely to \emph{rank} hyperparameter settings during the grid search.

\paragraph{Selected hyperparameters.}
From the grid search in Fig.~\ref{fig:hparam-heatmaps}, the best settings we use in the main results are:
\begin{table}[H]
\centering
\caption{Chosen $(\lambda, E_{\mathrm{attn}})$ from the hyperparameter search. Higher Optim Value is better.}
\label{tab:hparam-best}
\begin{tabular}{lccc}
\toprule
Dataset & $\lambda$ & $E_{\mathrm{attn}}$ & Optim Value \\
\midrule
DecoyMNIST & $8$ & $13$ & $0.1015$ \\
ColorMNIST & $160$ & $11$ & $0.1069$ \\
\bottomrule
\end{tabular}
\end{table}

\end{document}